\documentclass[letterpaper, 10 pt, conference]{ieeeconf}  

\IEEEoverridecommandlockouts                              

\usepackage{graphicx} 
\usepackage{amsmath} 
\usepackage{amssymb}  
\usepackage[linesnumbered,ruled,vlined]{algorithm2e}
\usepackage{booktabs}
\usepackage{cancel}
\usepackage{subcaption}

\newtheorem{remark}{Remark}
\newtheorem{assumption}{Assumption}

\newtheorem{problem}{Problem} 

\newtheorem{definition}{Definition}

\title{\LARGE \bf
Underwater Caging and Capture \\for Autonomous Underwater Vehicles
}

\author{\"Ozer \"Ozkahraman and Petter \"Ogren \\
KTH Royal Institute of Technology\\
\{ozero, petter\}@kth.se}

\begin{document}

\maketitle
\thispagestyle{empty}
\pagestyle{empty}

\begin{abstract}
In this paper, we consider the problem of caging and eventual capture of an underwater entity using multiple Autonomous Underwater Vehicles (AUVs) in a 3D water volume
We solve this problem both with and without taking bathymetry into account.
Our proposed algorithm for range-limited sensing in 3D environments captures a finite-speed entity based on sparse and irregular observations.
After an isolated initial sighting of the entity, the uncertainty of its whereabouts grows while deployment of the AUV system is underway.
To contain the entity, an initial cage, or barrier of sensing footprints, is created around the initial sighting, using islands and other terrain as part of the cage if available.
After the initial cage is established, the system waits for a second sighting, and the possible opportunity to create a smaller, shrinkable cage.
This process continues until at some point it is possible to create this smaller cage, resulting in capture, meaning the entity is sensed directly and continuously.
We present a set of algorithms for addressing the scenario above, and illustrate their performance on a set of examples.
The proposed algorithm is a combination of solutions to the min-cut problem, the set cover problem, the linear bottleneck assignment problem and the Thomson problem.
\end{abstract}

\section{INTRODUCTION}
\label{sec:intro}

Caging is the act of creating a connected perimeter around a given target such that the target cannot leave the area without breaching the perimeter, but is still free to  move within it \cite{rodriguez2012caging}. 
While the cage can be physical, like in grasping \cite{fink2008multi}, it can also be based on sensory detection where the target is simply detected but not obstructed when it tries to escape. 
In this paper we consider the detection case. The caging problem can be found in areas such as wildlife surveillance, escorting, security and herding \cite{varava2017herding}.

\begin{figure}[t]
 \centering
 	\includegraphics[width=.48\linewidth]{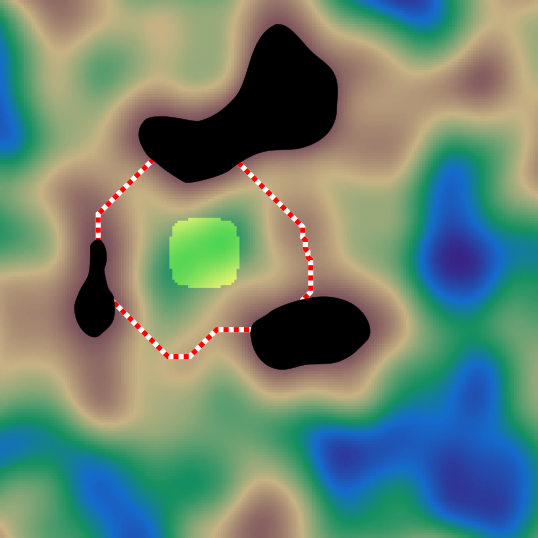}
    \includegraphics[width=.48\linewidth]{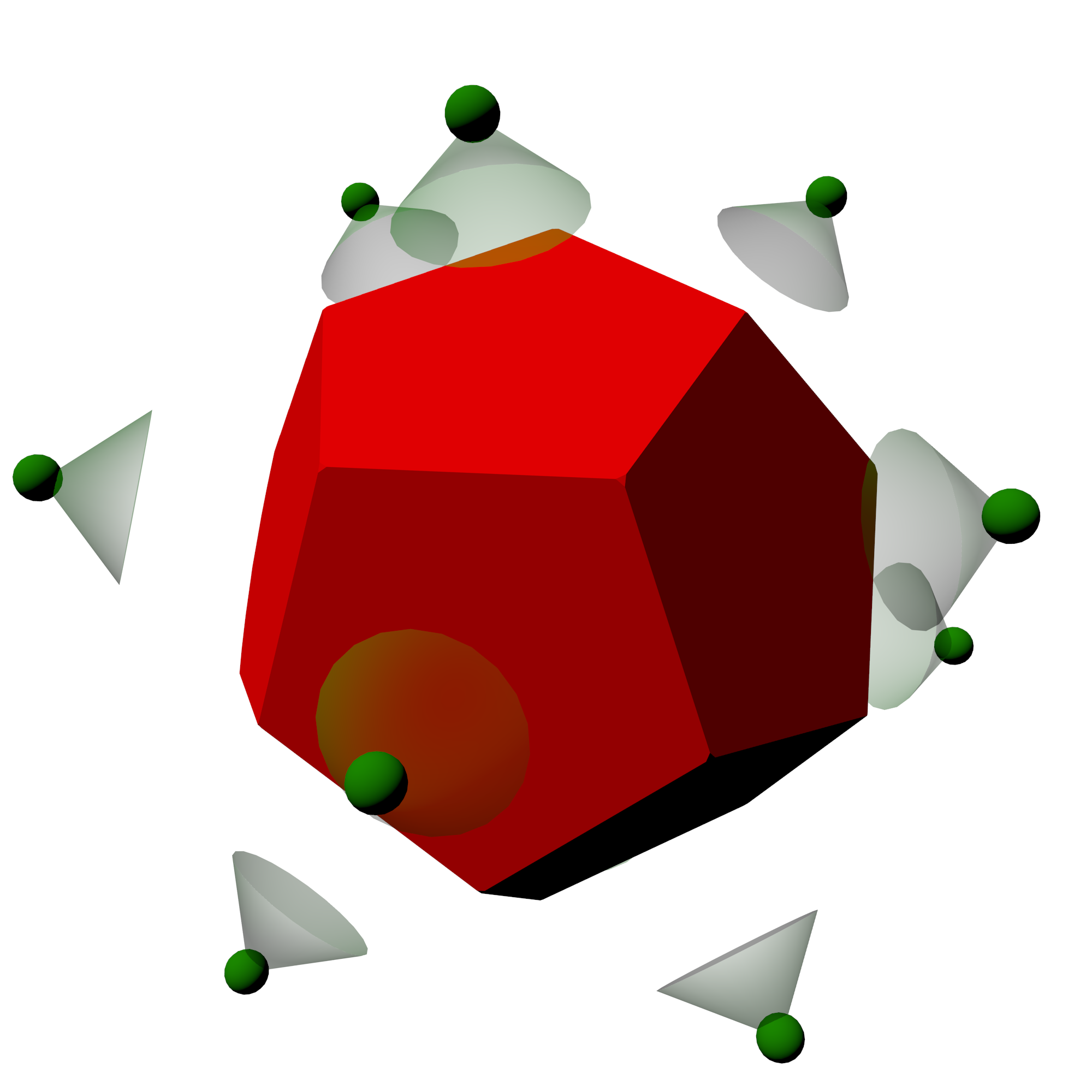}    
    \caption{
    A cage is established either with the help of islands and the seabed (left) or in the free water volume (right).
    The former is more resource efficient for large cages whereas the latter is independent of static obstacles and can be arbitrarily shrunk to guarantee capture.
    The AUV sensor footprints are hinted with the green cones and are not fully shown for clarity.
    }
 \label{fig:results}
\end{figure}

\begin{figure*}
\centering
\begin{minipage}[t]{.31\linewidth}
 	    \centering
        \includegraphics[width=\linewidth]{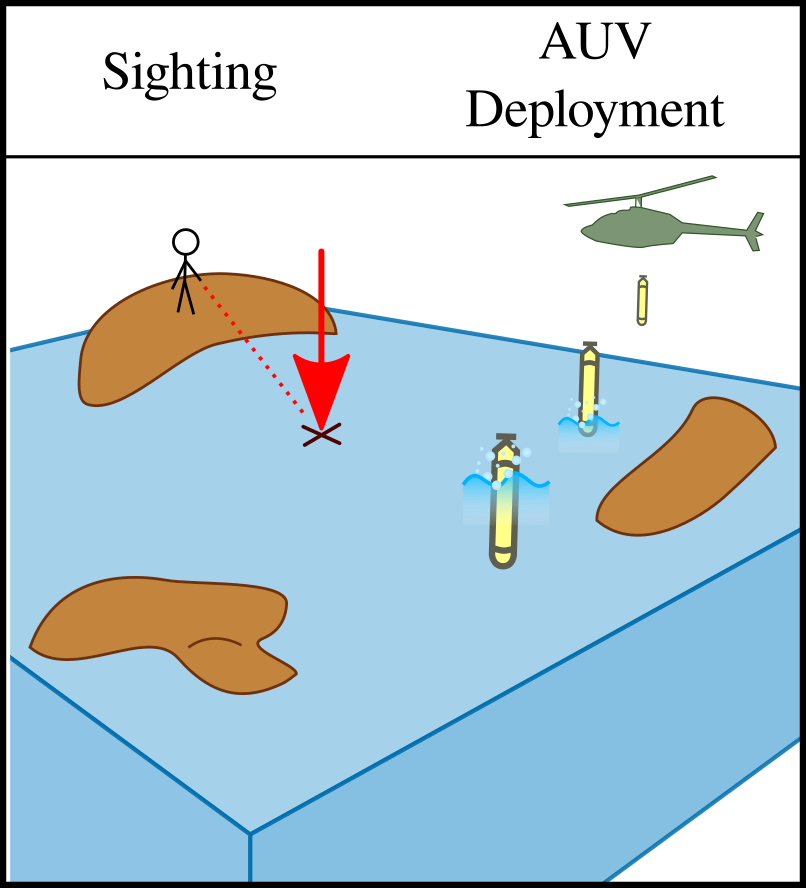}
 		\subcaption{Initial sighting and AUV deployment.}
 		\label{fig:overall_a}
    \end{minipage}
 	\begin{minipage}[t]{.31\linewidth}
 	    \centering
 		\includegraphics[width=\linewidth]{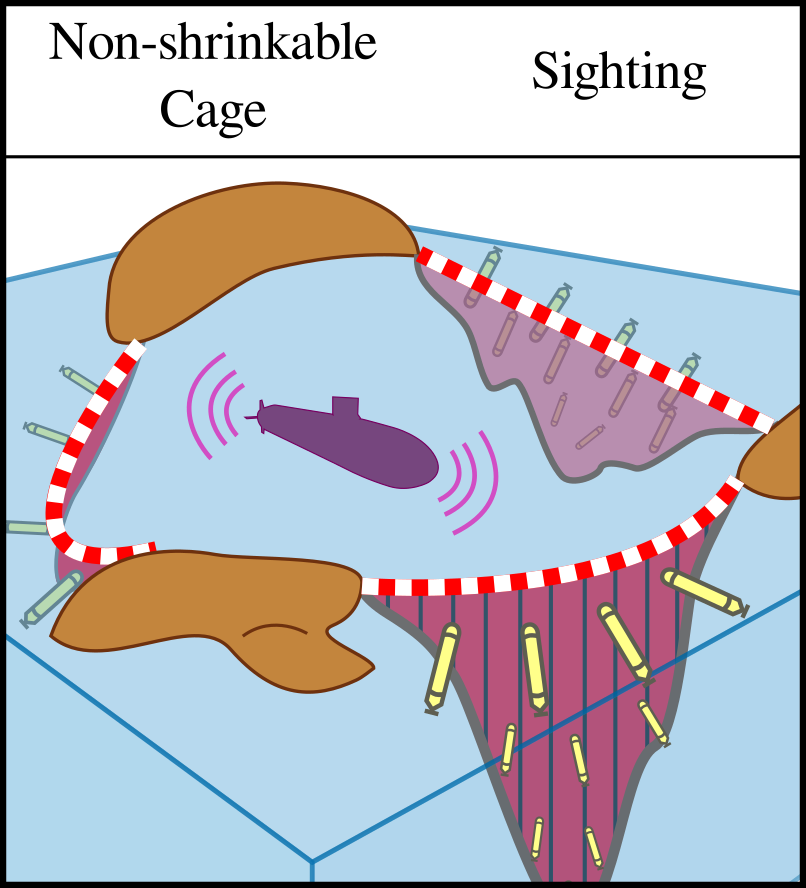}
 		\subcaption{Containing cage and secondary sightings.}
 		\label{fig:overall_b}
    \end{minipage}
    \begin{minipage}[t]{.31\linewidth}
 	    \centering
 		\includegraphics[width=\linewidth]{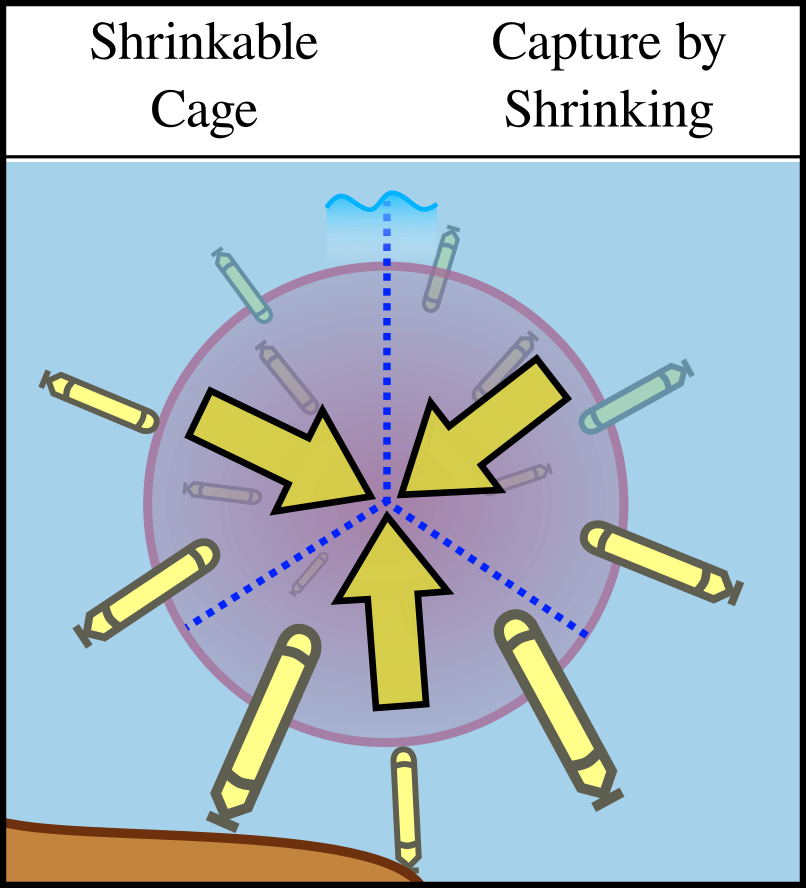}
 		\subcaption{Capturing cage followed by shrinking.}
 		\label{fig:overall_c}
    \end{minipage}
 \caption{Steps of the overall algorithm.}
 \label{fig:overall}
\end{figure*}

In this paper we consider a scenario where the entity is either an aquatic mammal such as a whale, or an intruding submarine or AUV, see Figure~\ref{fig:overall}. 
The scenario starts with the first sighting of an entity at a given location (Figure~\ref{fig:overall_a}). 
After this initial sighting, the entity might move with finite speed without further detection, thus an uncertainty region of possible locations starts growing around the last known location of it.
In order to guarantee the capture of the entity, the expansion of this uncertain region must be limited and eventually eliminated completely by autonomous underwater vehicles(AUVs).

We will now provide an informal overview of the algorithm.
To contain the entity a so-called \textit{containing cage} is formed, possibly making use of terrain such as islands and shallower parts of the water volume (Figure~\ref{fig:overall_b}).
Using the terrain at this stage allows for the creation of larger cages with fewer vehicles.
This is done by discretizing the problem into a graph, letting edge costs correspond to the area of the vertical cage segment between the edge on the surface, and the seabed right underneath it. 
Then the needed cage segments are found by solving a min-cut problem in this graph.
AUV positions are found by solving a set cover problem on these vertical segments, after which the AUVs are assigned their positions using the linear bottleneck assignment problem.
Once the positions for all AUVs are known and assigned, the AUVs are moved to create the containing cage. 
This containing cage is kept stable until a second momentary detection happens. 
This second sighting is expected to be from the AUVs that are already in the containing cage formation since they are now closer to the entity and can be equipped with more accurate sensors.
After the new detection, a new uncertainty region starts to grow centered on the new sighting location.
The AUVs are much closer to the sighting location this time, and the possibility for a spherical so-called \textit{capturing cage} is explored (Figure~\ref{fig:overall_c}), by solving a Thomson problem and a linear bottleneck assignment problem. 
If such a cage can be formed with the available AUVs, before the uncertainty region grows beyond the physical limitations of the AUVs, the containing cage is replaced by the spherical capturing cage. 
Finally, this spherical cage is uniformly shrunk until capture, i.e., accurate and continuous detection of the entity by the AUVs own range limited sensors. 

The main contribution of this paper is a two-step planning algorithm for underwater caging and capture that first contains the entity within a large volume and then captures it by guaranteeing that an AUV will be able to get within continuous sensing range.  
The algorithm takes into account the bounded sensing range and movement speed of the AUVs and tries to minimize the number of AUVs needed within these constraints.
To the best of our knowledge, no such algorithm is described in the literature.

The outline of this paper is as follows, first, in Section~\ref{sec:related_work} we describe related work followed by the relevant background information in Section~\ref{sec:background}. Then, in Section~\ref{sec:problem}
we formalize the problem.
In Section~\ref{sec:solution} we then describe the two steps of the solution, and illustrate the approach in Section~\ref{sec:simulation_results}.
Conclusions are found in Section~\ref{sec:conclusions}.

\section{RELATED WORK}
\label{sec:related_work}

The problem addressed in this paper is related to the so-called
Pursuit-evasion problem, defined as game where a group of pursuers attempt to guarantee the capture of an evader.
The vast majority of the work on pursuit-evasion, has considered 2D environments, as described in recent surveys, \cite{Chung2011, Robin2016}, and 3D pursuit evasion has largely been an open problem.

In \cite{gerkey2006visibility} the authors show that for 2D problems with polygonal obstacles and arbitrarily fast evaders, it is NP-hard to find the minimal number of pursuers needed to guarantee capture.
In our case, the evading entity has finite speed and the problem is set in 3D.
In addition, the entity in our problem is assumed to be unaware of our capture attempt, thus it is not moving actively to avoid our AUVs.

In a recent work \cite{Franchi2016}, the authors present a control framework for encircling a moving target in 3D. 
Encircling is an adjacent method to caging, where an incomplete cage is in constant motion and the captor speed compared to the evader speed makes the cage possible. 
The captors create a circle on a plane around the entity and use that plane as their basis of movement. 
While the 2D plane rotates and moves in 3D, the encircling agents are always on that plane thus the entity is never completely caged.
In their proposed future work, the authors mention encircling on multiple planes. 
In contrast, our work creates a complete 3D cage at all times around the target, removing the requirement of sensing the entity continuously and thus it can be used with range-limited sensors. 
A similar encirclement problem is addressed in \cite{Yao2017} and solved with a distributed control law for single-integrator agents. 
They consider the case where any agent only knows the position of the two neighboring agents on a circle. 
Both of these works rely on the entity being slow and large enough that the encirclement can catch it before it fully escapes. 
In this paper, we need the entity to have bounded speed only until the capturing cage is established, afterwards we guarantee containment and capture regardless of its speed.

The authors of \cite{Aranda2014} present a method of control in 3D to cage a target using multiple agents. 
In contrast to us, they assume that the 3D shape to create with the vehicles is given externally. 
Unlike \cite{Franchi2016}, this shape is not required to be circles on planes. 
The focus of the work is on keeping this given formation while allowing movement rather than generating the formation. 
The methods described in \cite{Aranda2014} can thus be used in combination with the methods described in this paper where the methods proposed here generate the shapes that \cite{Aranda2014} requires as input.

In \cite{bhadauria2012}, the authors consider the problem of pursuit-evasion inside a closed and fully observable 2D polygonal space. 
They assume the captors always know the position of the evader and vice versa. 
They use identical captors and evaders in terms of movement capabilities. 
Their main contribution is that they show that three captors are always enough and sometimes necessary to guarantee capture for polygonal worlds with holes.
The biggest difference between \cite{bhadauria2012} and ours is the knowledge both the AUVs and the entity have of each other. 
The assumption of continuous knowledge of entity position is very unlikely to hold in an underwater setting since sensing and communication ranges are limited. 
The authors of \cite{bhadauria2012} also consider the problem in 2D.

The work done in \cite{Reed2016} use experiments to evaluate the capability of acoustic communication underwater. 
They show that recent advances  in acoustic underwater communications enables a dynamic multi-agent system such as the one described in this paper. 
Communication is useful for our work in the sense that the AUVs in our work need to communicate in order to coordinate inbetween the phases of operation, and overall performance is thus enhanced by more reliable and frequent communications. 

In this paper, we go beyond the works described above and propose a two-step algorithm for caging and capture.
Our method creates a plan that first restricts the entity to a known volume using a near minimum number of agents, followed by the creation of a secondary cage that shrinks that volume to guarantee the capture of the entity. 
We make minimal assumptions about the environment in which the operation takes place and we incorporate the sensing capabilities of the agents into the plan.

\section{Background}
\label{sec:background}
In this section we will review four problems from the literature that we will use as part of the proposed solution. 
First the Linear Bottleneck Assignment Problem (LBAP), then the Thomson Problem (TP) and then Max-flow Min-cut Problem (MCP).
Finally we look at the Set Covering Problem (SCP).

\begin{problem}  \emph{Linear Bottleneck Assignment Problem (LBAP) } \newline
 Given two sets, $A,T$ and a cost function $C:A \times T \rightarrow \mathbb{R}$
 find a bijection $f:A \rightarrow T$ that solves
 $$
 \min_f \max_{a \in A} C(a,f(a)).
 $$
 This problem is solvable in polynomial time, $\mathcal{O}(\max(A,T)^2)$, see  \cite{ARMSTRONG1992179}.
\end{problem}
This problem is used for pairing AUVs to positions such that the time to completion of the longest path is minimized.
This requirement comes from the fact that the cage is complete only when all AUVs are at their desired positions. 
Before that, there will be a hole in the cage.

\begin{problem}  \emph{Thomson Problem (TP) } \newline
 The problem of regular distribution of points around a sphere is known as the Thomson Problem,~\cite{Thomson1904}. 
 \end{problem}

The Thomson Problem was originally proposed in terms of trying to minimize the electric potential energy of a system of electric charges moving on a sphere. 
These charges push each other with a force proportional to $1/l_{ij}^2$ where $l_{ij}$ is the distance between~the charges $i$ and $j$.
It has also been shown, that apart from the so-called Platonic solids (PS) \cite{pugh1976polyhedra}  with $N=4,6,12$ it is not possible to find a single distance $l$ such that $l_{ij} = l$ for all $i,j$ given they are neighbors, \cite{Wales2006}.
To find approximate solutions to the Thomson Problem we will simulate $N$ electric charges on a sphere of unit radius. 
This problem is useful for us due to the fact that our captors are homogeneous in terms of sensing footprint.
Since the sensing footprints are the same, the distances between the AUVs must also be equal, or as close to equal as possible.
Later we will show that it is not always possible to find such equal distributions and that the resulting cage becomes less efficient because of this.

\begin{problem}  \emph{Max-flow Min-cut  Problem (MCP)} \newline
Given a graph $G=(V,E)$ with an edge cost $C:E \rightarrow \mathbb{R}_+$ and two subsets of vertices $v_1,v_2 \subset V$ such that $v_1 \cap v_1 = \emptyset$.
Find a set of edges $E_{cut}\subset E$, of minimal aggregated cost $\Sigma_{e \in E_{cut}}C(e)$ such that the graph $(V, E \setminus E_{cut})$ contains no path between $v_1$ and $v_2$.

This problem can be solved in polynomial time, with complexity $\mathcal{O}(E^2 V)$, \cite{edmonds1972theoretical} or $\mathcal{O}(VE+V^2 logV)$, \cite{stoer1997simple}. 
The reason it is called max-flow min-cut is that the min-cut formulation above is actually equivalent to a problem where the costs are flow capacities and one wants to know how much flow can pass between $v_1$ and $v_2$.
\end{problem}
This problem formalizes our requirement to reduce the number of AUVs required to create a large containing cage.
The solutions to this problem effectively minimizes the required fleet size and uses islands as part of the containing cage when it is advantageous to do so.

\begin{problem} \emph{Set Cover Problem (SCP)} \newline
Given a set $U$ and a collection of subsets $S_i \subset U$, $S=\{S_i\}$ such that the union $\cup S_i = U$.
Find the smallest subset  $C \subset S$ such that $\cup_C S_i = U$.
The general set cover problem is NP-hard \cite{korte2012combinatorial} but greedy algorithms can find approximate solutions in polynomial time \cite{slavik1997tight}. 
In special cases, where $S_i$ are disc shaped, the so-called discrete unit disc cover problem can be solved in near-linear time \cite{agarwal2014near}.
\end{problem}
We make use of this problem in order to find exactly where each of our captors should be positioned such that there are no holes in the containing cage.
The walls of the containing cage(the set $U$) must be covered by the AUV sensors (the discs $S_i$) such that there is no point in $U$ that is not covered.
Note that overlapping discs are allowed and are in fact required for full coverage.

With this background we are ready to formulate the main problem.

\section{Problem Formulation}
\label{sec:problem}

Given an entity with position $e(t) \in \mathbb{R}^3$ and a set of AUVs with positions $p_i(t) \in \mathbb{R}^3, i= 1 \ldots N$ and orientations $R_i \in SO(3)$, we want to move the AUVs such that the entity is eventually sensed continuously by at least one of the AUVs, in terms of the sensing region described below. 
When this happens the mission is successful and our method ends.
Both the entity and the AUVs are constrained to move in the obstacle free space $\mathcal{F} \subset  \mathbb{R}^3$. 
This is typically the water volume bounded by the seabed, the surface and a set of islands. 
Let $\delta \mathcal{F}$ denote the boundary of $\mathcal{F}$. 
The maximum speeds are given by $v_e$ and $v_p$ respectively.

Each AUV has a sensing region $s(p_i, R_i) \subset  \mathcal{F}$ in the shape of a sphere, or a cone.
More complex shapes could also be relevant, depending on occlusions in  $\mathcal{F}$ or other factors limiting sensor range such as salinity and water temperature.
Most such factors are unknown at the planning stage, so we model the sensor as either a sphere or a cone and set the sizes of these footprints conservatively to account for the unknown factors.

\begin{assumption} \emph{Irregular sightings} \newline
 At irregular time instants $t = t_k, k=1, \ldots , K$ we are given the position of the entity $e(t_k)$. 
 This corresponds to the entity doing some form of motion, sensing, or communication transmission that enables us to temporarily localize it (Figure~\ref{fig:overall}a,b). 
 We furthermore assume that these sightings will appear more often if we have a better idea of the possible whereabouts of the entity in terms of a smaller contaminated set, see below.
\end{assumption}

\begin{definition}\label{def:contaminated}
 \emph{Contaminated Set} \newline
Given only the knowledge from the irregular sighting, let the contaminated set $\mathcal{C}(t) \subset \mathcal{F}$ capture the knowledge of the entity available to the AUVs. 
At the times of sighting, $t = t_k, k=1, \ldots, K$, we have that $\mathcal{C}(t_k)=e(t_k)$. 
In between the sightings the set grows with the maximal entity speed $v_e$. 
Thus $\mathcal{C}(t) \subset \{x : ||x - e(t_k) ||\leq v_e (t-t_k) \} \forall k, t \geq t_k$.
Note that $\mathcal{C}(t)$ is thus bounded by a sphere, but it is not always equal to a sphere since it is also bounded by both  $\delta \mathcal{F}$ and $\cup_i s(p_i,R_i)$.
\end{definition}

\begin{definition} \emph{Caging Formation} \newline
Given a contaminated region $\mathcal{C}(t)$, we say that the AUV poses 
$p_i(t),R_i(t)$ are in a Caging Formation if the union of their sensing volumes $\cup_i s(p_i, R_i)$ partitions $\mathcal{F}$ into disconnected components out of which only one contains $\mathcal{C}(t)$.
\end{definition}
Note that as long as the Caging Formation is kept, the contaminated region $\mathcal{C}(t)$ cannot expand outside of this single connected component of $\mathcal{F}$. 
Thus the formation cages the contaminated region and the entity within. 
For examples of different types of caging formations see Figure~\ref{fig:results}.

\begin{definition} \emph{Shrinkable Caging Formation} \newline
A set of continuous AUV trajectories $p_i(t)\in \mathcal{F}, R_i(t)$ for $t \in [t_0,t_f]$  such that for each $t \in [t_0,t_f)$, $p_i(t)$ are  in a Caging Formation, and
the volume of $C(t)$ tends to $0$ as $t \rightarrow t_f$.
\end{definition}

Note that if the AUVs are in a shrinkable caging formation we just need to execute the corresponding  trajectories and thereby shrink the contaminated set to the empty set, and thus guarantee the intersection of the entity position with at least one of the sensing sets, $e(t') \in s(p_j(t'),R_j(t'))$ for some time $t'$ and AUV $j$.

\section{Proposed Solution}
\label{sec:solution}

The overall proposed solution is described in Algorithm~\ref{alg:overall}.

\begin{figure}[ht]
 \centering
 	\includegraphics[width=.30\linewidth]{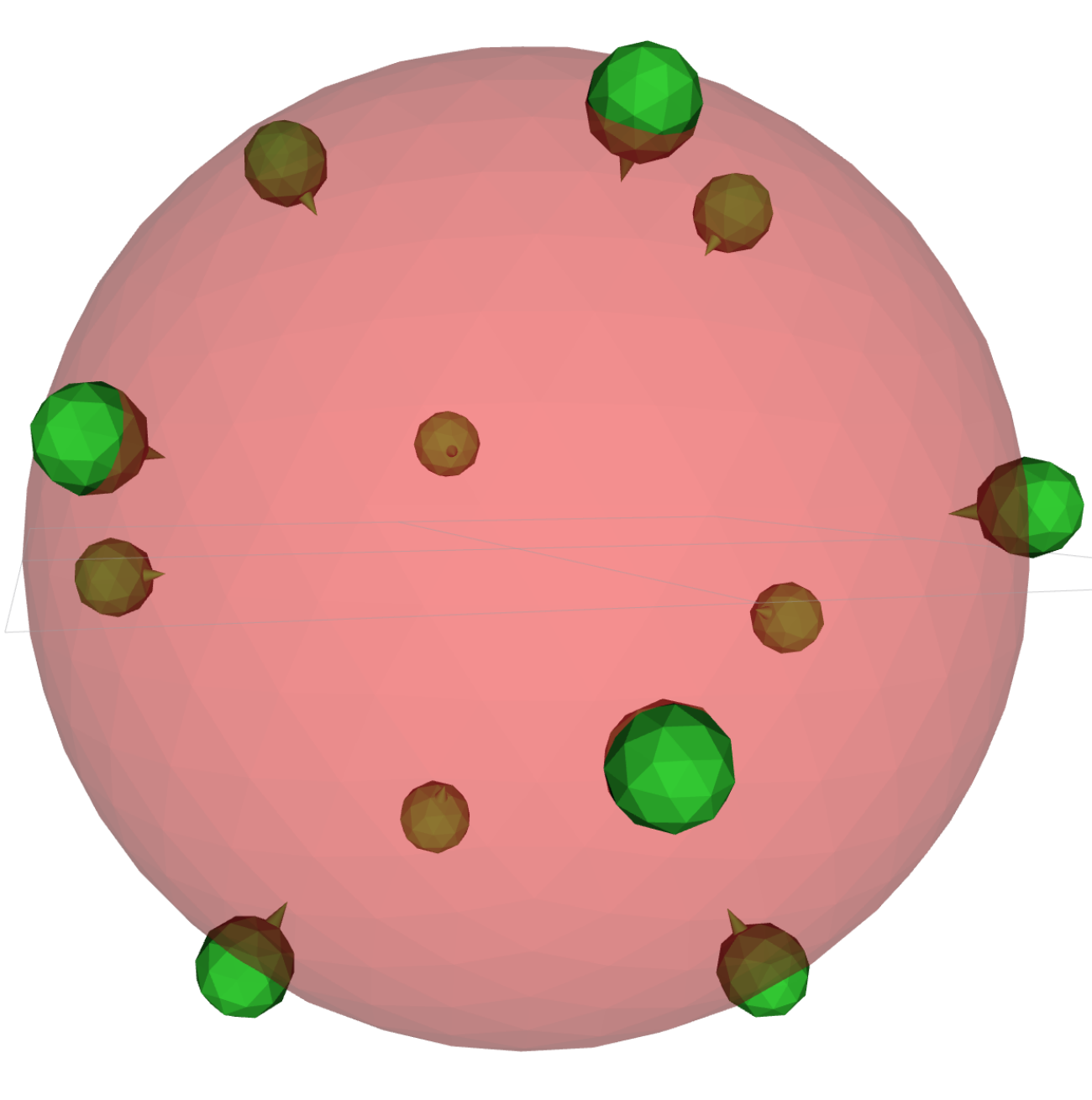}
    \includegraphics[width=.30\linewidth]{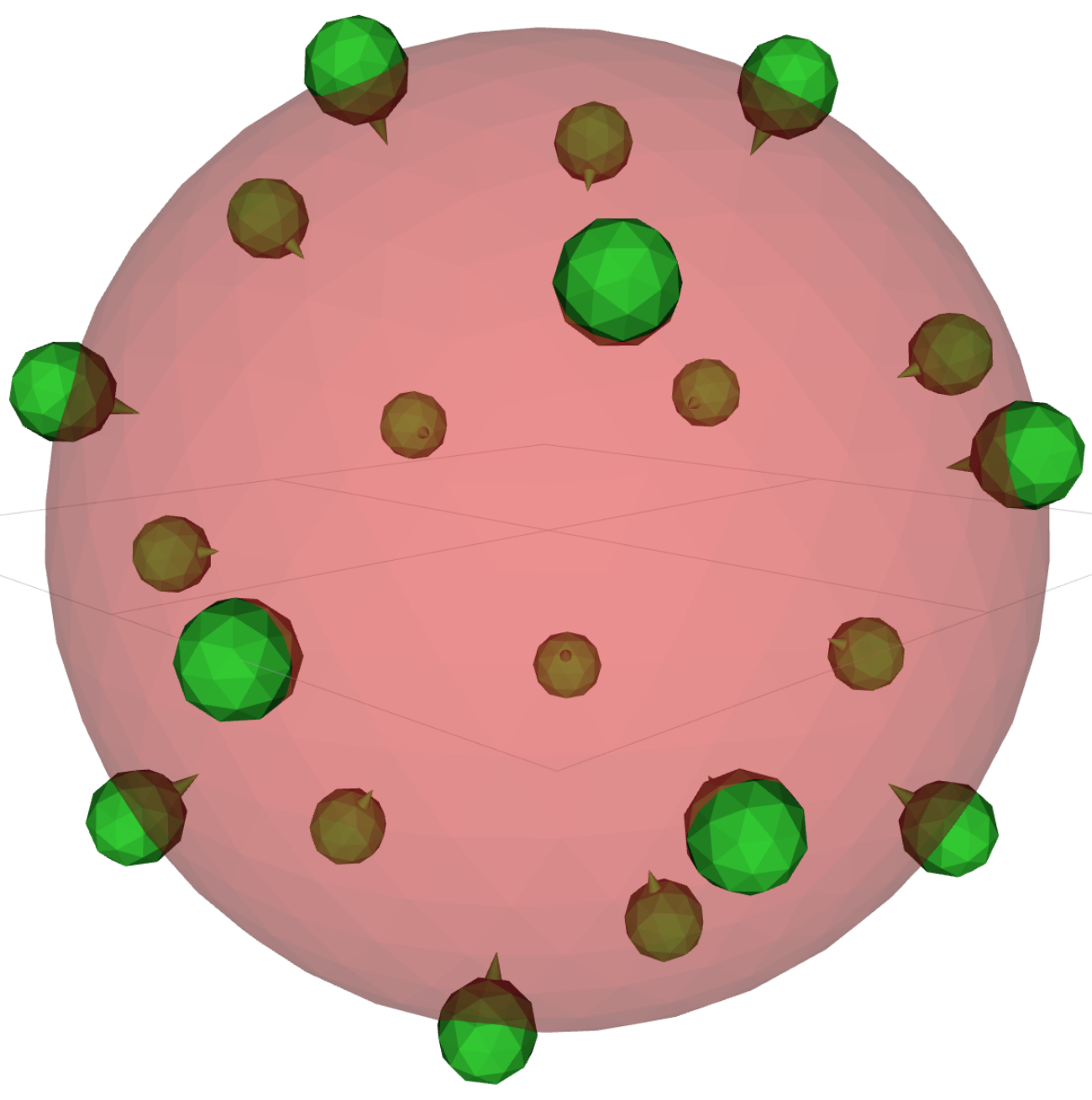}
    \includegraphics[width=.30\linewidth]{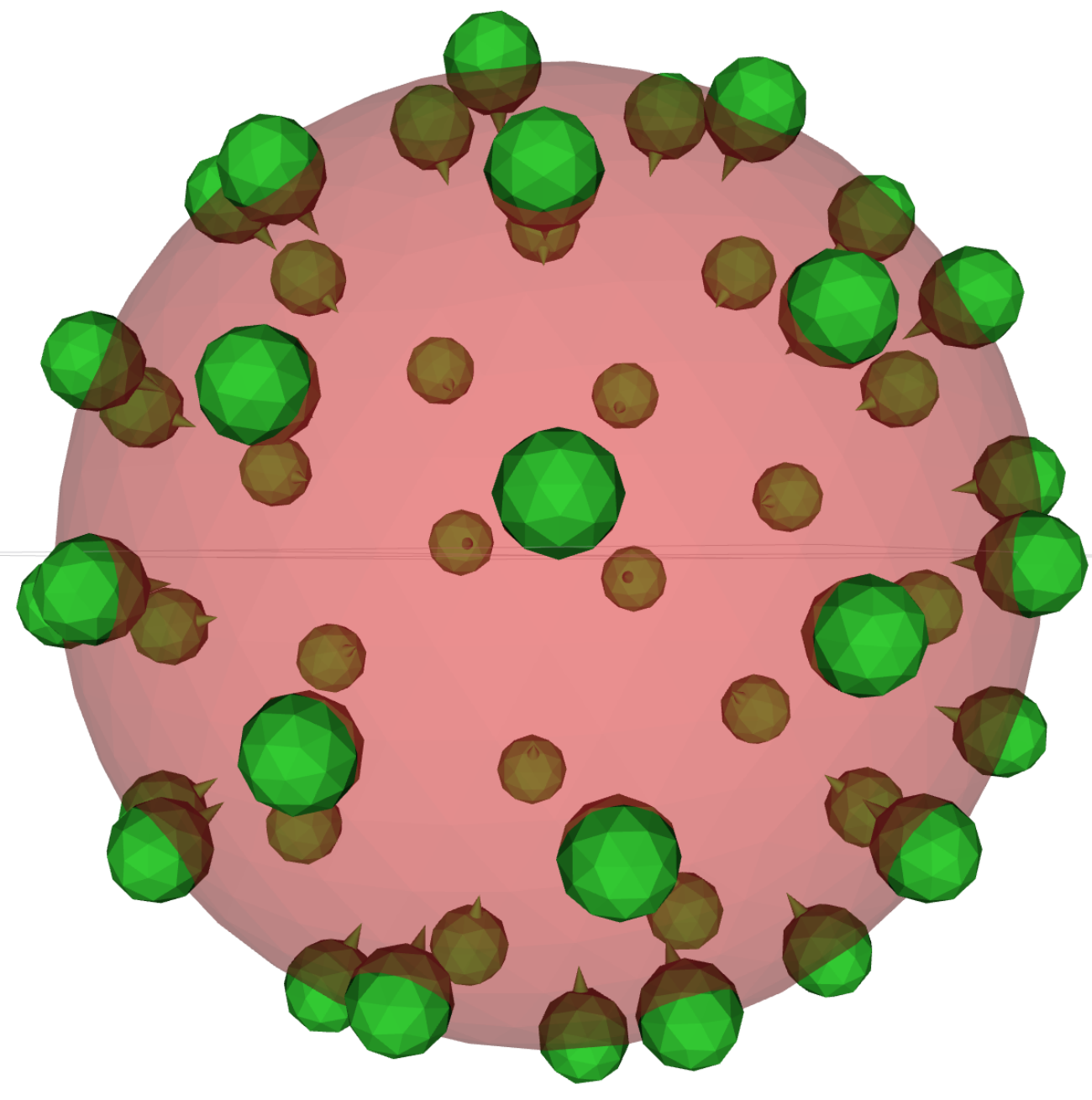}
    \caption{Shrinkable cage positions for 12, 20, 50 AUVs(green) on the unit sphere(red). The full sensor ranges are not shown for visual clarity.}
 \label{fig:cages}
\end{figure}

\begin{algorithm}[!h]
  \caption{Cage and Capture (execute at each new entity sighting}
  \label{alg:overall}

\KwIn{Time and position of latest entity sighting $e(t_k)$}
\KwIn{Current positions of the AUVs $p_i(t)$}

\KwResult{AUV trajectories to be executed}

  \If{Exists Reachable Shrinkable Caging Formation }{
    	Go To Formation \;
    	Shrink \;
    }
    \ElseIf{ Exists Reachable Formation with smaller contained volume than current}{
    	Go To Formation \;
	Wait for next sighting \;
    }
\Else{  Wait for next sighting}
\end{algorithm}

\subsection{Reachable Cages}

Given that the contaminated set grows steadily after each sighting, a set of candidate positions represent a cage only for a given time, after which the contaminated set is no longer in only one part of the free space.
Thus, given a set of initial vehicle positions, candidate caging positions, and the contaminated set, we say that the candidate caging positions are a Reachable Cage if there is an assignment of the caging positions to the vehicles such that all of them can reach their destinations fast enough. 
This problem is an instance of the Linear Bottleneck Assignment Problem (LBAP), see Section \ref{sec:background}, to which there exists polynomial time solutions.
To conclude, given a cage we can solve an instance of the LBAP to decide if it is a reachable cage.

\subsection{Finding Non-Shrinkable Cages}
\begin{figure*}[t]
 \centering
 	\includegraphics[width=0.19\linewidth]{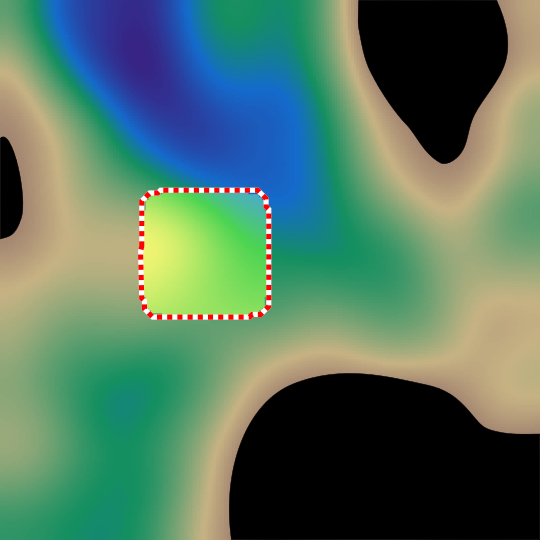}
 	\includegraphics[width=0.19\linewidth]{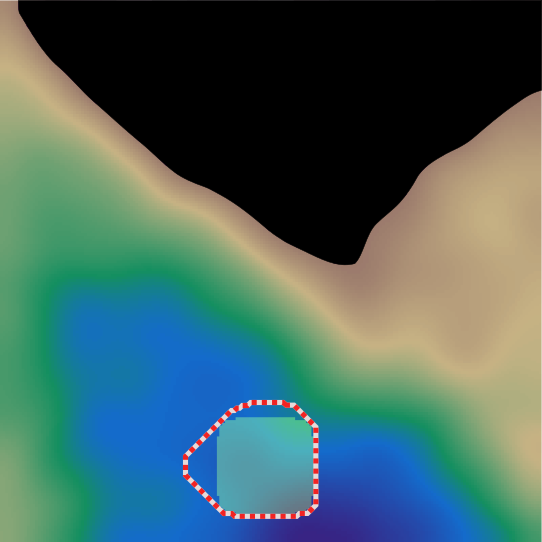}
    \includegraphics[width=0.19\linewidth]{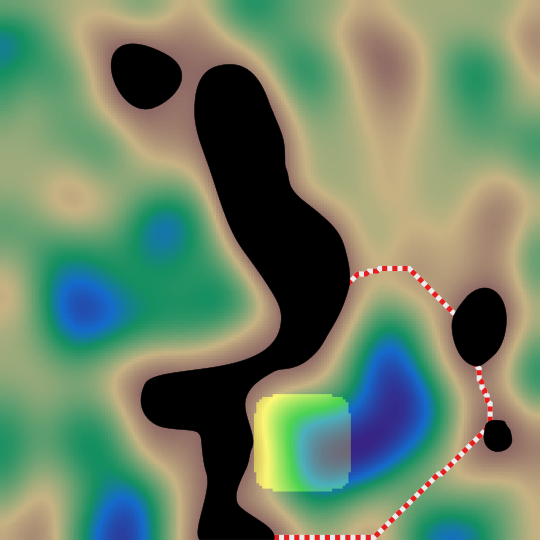}
    \includegraphics[width=0.19\linewidth]{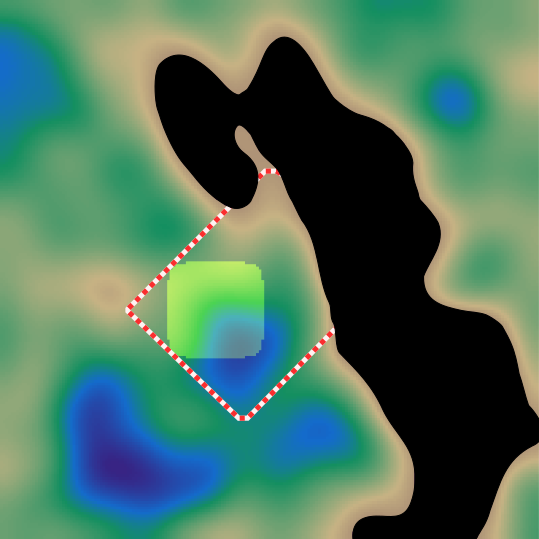}
    \includegraphics[width=0.19\linewidth]{non_shrink_5.png}
 \caption{Some non-shrinkable, containing cages seen from the top. Bright yellow area shows the contaminated area $F$ and the dashed line around it shows the cage found. Island areas are shown in solid black. It can be seen that the cage uses different amounts of islands and shallow areas in order to optimize total vertical surface area. In the rightmost figure, the cage exploits three islands while on the leftmost, the cage is using no islands and is tightly wrapped around $F$.}
 \label{fig:curve_results}
\end{figure*}

A cage can be formed by a combination of the static free-space boundary $\delta \mathcal{F}$ and the AUV positions $p_i(t)$. 
Exploiting the geometry of $\delta \mathcal{F}$ can enable us to create very efficient (in terms of AUVs required per volume caged), but not shrinkable, cages, see Figure \ref{fig:curve_results}. 

When most parts of a cage is made up of the boundary of the free space $\delta \mathcal{F}$, such as the seabed, the surface and islands, explicit account must be taken to those features.
When this is the case it is furthermore reasonable to think that the contaminated area is mainly spreading in the horizontal direction, having already reached the surface as well as the seabed above and below the point of sighting. 
This shape is essentially a very large sphere that is cut on the top and bottom by planes and resembles a cylinder more than a sphere at this point. 
With most of the spreading taking place in the horizontal direction, it would make sense to make the cage walls vertical. 
Although we do not explicitly take AUV dynamics into account in this paper, we note that vertical walls are in fact easier to realize with most AUVs with forward looking conical sensors.
Therefore we will now consider the problem of finding vertical barriers around the contaminated volume.

\begin{problem}
 Assume we are given a graph discretization of the horizontal plane $G=(V,E)$. 
 Let $v_c$ be the set of vertices inside the contaminated region, $v_c=\{v \in V: v \in C(t)\}$.
 In this  graph we assign edge costs to all edges that are proportional to the edge length times the average depth under that edge,  $c:E \rightarrow \mathbb{R}_+$ as $c((v_i,v_j))=||v_i - v_j|| (D_m(v_i)+D_m(v_j))/2$, i.e., the area of a barrier segment from surface to seabed under that edge, where $D_m(v)$ is the depth at vertex $v$. 
 Any closed loop around the contaminated set $v_c$ would now correspond to a cage and the aggregated cost of that loop would correspond to the total surface area of that cage.
\end{problem}
To solve the problem above we use Algorithm~\ref{alg:vertical}, where we use the Min-cut Problem, followed by the Set Covering  Problem, see Section~\ref{sec:background}. 
Then, we use LBAP as mentioned in the previous section to check if this cage is reachable.

\begin{algorithm}[h!]
\KwIn{Number and state of AUVs, sensing regions, and contaminated set: $N, p_i, R_i, s(p_i,R_i), C(t)$}
\KwResult{AUV destinations $[f_1 \ldots f_N] \in \mathbb{R}^3$  }
Create a planar graph $G=(V,E) $\;
Create the dual $G'$ of $G$ and let the costs of the new edges be the same as the cost of the old edges they cross\;
Let $v_c$ of the new graph be the vertices corresponding to faces inside $C(t)$ and $v_b$ be the vertex corresponding to the face outside the planar graph $G$ (escaping from the map)\;
Solve the Min-cut problem between $v_c$ and $v_b$\;

This cut is the optimal cage\; 
Remove edges with zero cost (on land)\;
Solve SCP to position AUVs on vertical barrier segments \;
Solve LBAP to assign AUVs to the desired positions, $f_i$, to cover the vertical barrier segments corresponding to the optimal cage\;
\Return{$[f_1 \ldots f_N]$}
    \caption{Non-shrinkable Vertical Cage}
    \label{alg:vertical}
\end{algorithm}

\subsection{Finding Shrinkable Cages}
In general, for a cage to be shrinkable it needs to depend less on the unchanging terrain boundaries $\delta \mathcal{F}$ and more on the controllable AUV locations  $p_i(t)$. 
This is due to the fact that if most of the cage is made of uncontrolled parts, then the number of controlled parts might not be enough to make a new cage without the uncontrolled parts.
When $\delta \mathcal{F}$ is not sufficiently exploitable, and given the fact that the contaminated set grows like a sphere, see Definition \ref{def:contaminated}, it is natural to look for spherical cage formations in order to tightly restrain the contamination while using a small number of AUVs.
Note that the spherical shape becomes important when the diameter of the sphere is less than the seabed to surface distance.
This corresponds to either very deep, open waters or small spheres in shallow areas.

\begin{assumption}
 Assume the sensor coverage volumes $s(p_i,R_i)$ are either spherical with radius $r_s$ or conical with some height $h$ and bottom radius $r_s$. 
 In either case $s_i$ includes a disc of radius $r_s$ which we will use to build the cage. 
 To simplify the presentation we let $p_i$ denote the center of the disc in both cases and compute the actual positions of the AUVs (a distance $h$ above the disc) in a post processing step for the conical case.
\end{assumption}
We now face the following problem

\begin{problem}
\label{prob:sphere}
 Compute an approximately spherical surface without holes from a set of $N$ discs with radius $r_s$.
\end{problem}

The pseudo-code for solving Problem~\ref{prob:sphere} is found in  Algorithm~\ref{alg:charges}.
First we approximately solve a Thomson Problem, see Section \ref{sec:background}, and then re-scale the solution to make sure the sensor coverage is complete.

\begin{algorithm}[h!]
\KwIn{$N, r_s$}
\KwResult{AUV destinations $[f_1 \ldots f_N] \in \mathbb{R}^3$  }
Create randomly positioned particles $[p'_1...p'_N]$ on the unit sphere $S_0$\;
Simulate damped particles moving on $S_0$ under repulsive Coulomb forces $f_{ij} \propto  \frac{p'_i-p'_j}{||p'_i-p'_j ||^3}$ until an equilibrium is reached\;
$edges = \mbox{Delaunay}([p'_1...p'_N])$\;
$l_{max} = \mbox{Maxlength}(edges)$\;
$l_{} = \sqrt{3} r_s$\;
$f_i \gets (l/l_{max}) p'_i ~~\forall i \in  1...N$\;
Solve LBAP to assign AUVs to the desired positions, $f_i$, to cover the disc shaped barrier segments corresponding to the optimal cage\;
\Return{$[f_1 \ldots f_N]$}
    \caption{Approximately Spherical Cage}
    \label{alg:charges}
\end{algorithm}

\begin{figure}[ht]
 \centering
 	\includegraphics[width=0.6\linewidth]{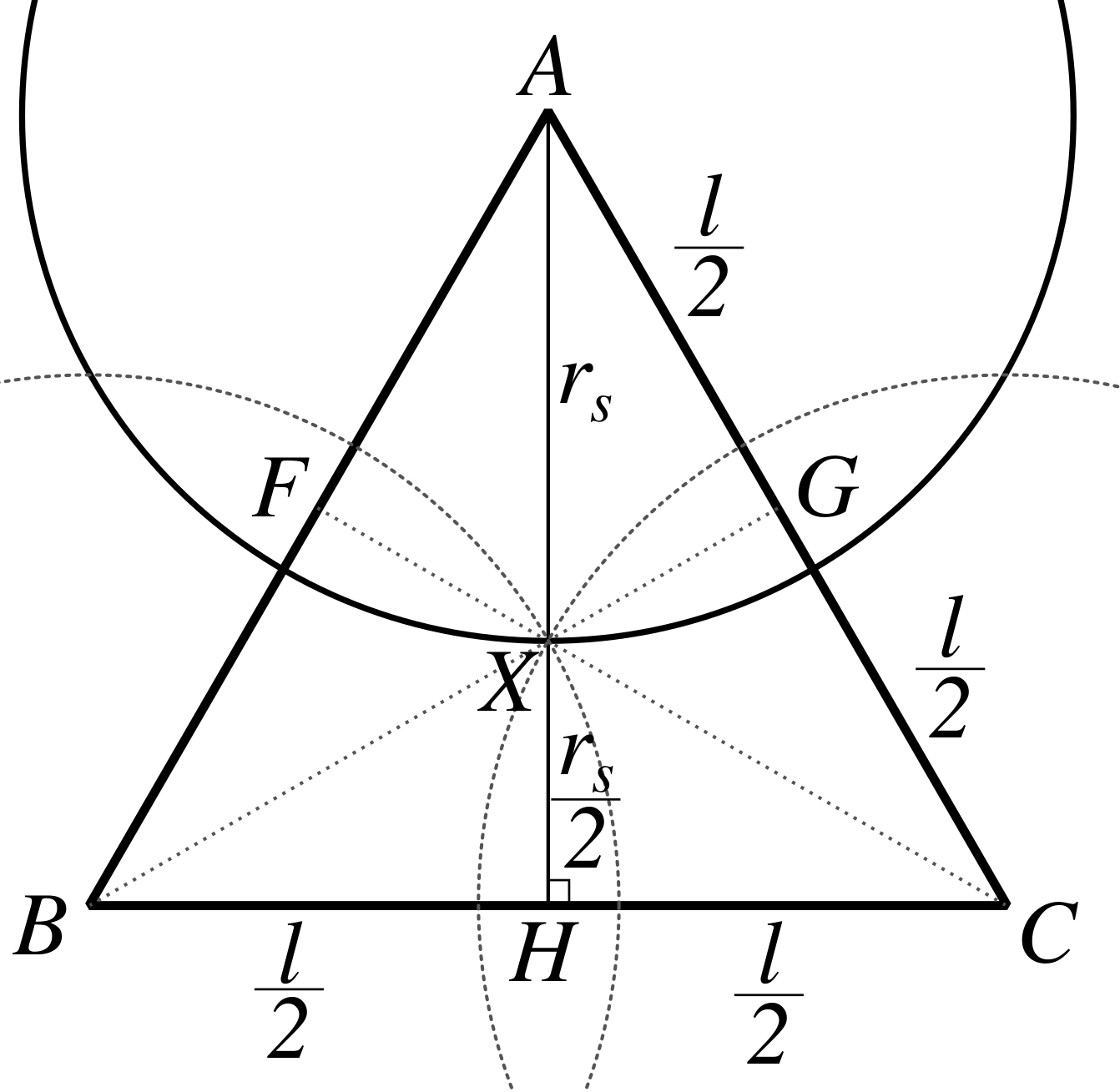}
 \caption{Three adjacent AUVs $A,B,C$ must be at most $l = \sqrt{3}r_s$ apart to make sure there is no sensor gap between them. Points $A,B,C$ are centers of the sensing surfaces with radius $r_s$. $\widehat{ABC}$ is an equilateral triangle of side $l$. Point $X$ is the intersection of the three medians ($|AH|$, $|BG|$ and $|CF|$) of the triangle. 
 Trigonometry now gives that $l = \sqrt{3}r_s$. }
 \label{fig:triangles}
\end{figure}

In lines 1-2 of the algorithm we find generic AUV positions on the unit sphere. It will be verified in Table~\ref{tab:dists} that when $N=4,6,12$ this approach does indeed recreate the classical Platonic solids. 
In line 3, Delaunay()  \cite{Delaunay1934}, is used to compute a graph where neighbouring positions are joined by edges, then in line 4, the maximal length of those edges is computed.
It is shown in Figure  \ref{fig:triangles} that keeping the edge lengths of the cage below or equal to $l=\sqrt{3}r_s$ guarantees that there are no holes in the sensor coverage. 
Thus in line 5-6 the formation is scaled to meet this requirement.

\begin{remark}
We note that when the edge lengths of the spherical cage are not equal, this is due to there being no triangulation of a sphere with $N$ many vertices with equilateral triangles.
In such cases, the non-equal edges become the dominating length when scaling the whole cage, leading to \textit{most} edges being unnecessarily short to accommodate the few long edges.
In order to keep our guarantee of full caging, this is necessary.
\end{remark}

\section{Simulation Results}
\label{sec:simulation_results}
In this section we will investigate and illustrate the performance of the different algorithms.

\subsection{Results on Non-shrinkable Cages}

In order to test our algorithm, we have generated random depth maps and random contaminated areas. 
The depth maps were generated using smoothed semi-structured noise and the contaminated areas were generated with smoothed random rectangles. 
The exact shape of the contaminated area makes no difference to our algorithm, so it was left the same for the experiments.
The depth maps were generated such that any point with value $0$ corresponds to an island and larger values correspond to the depth at that point. 
See Figure~\ref{fig:curve_results} for some of the results. 
The graph $G$ was generated such that every pixel in the depthmap is one vertex in $G$.

As can be seen from Figure~\ref{fig:curve_results}, 
the optimal barrier often, but not always, contains the $\delta \mathcal{F}$ of one or more islands.
In cases where islands are not present or are too far away from the contamination, our method finds the optimal cage closer to the contaminated area, using less or none of $\delta \mathcal{F}$.

\subsection{Results on Spherical Cages}
\label{sec:simulation_results_2}
In order to test our algorithm, we have generated caging formations for a fixed set of parameters where we could calculate the optimal inter-agent distances analytically and compared our results. 
Since our method starts in a randomized state, we have simulated many runs of the same problems to see the variation of the results.

\begin{figure}[htb!]
 \centering
 	\includegraphics[width=.99\linewidth]{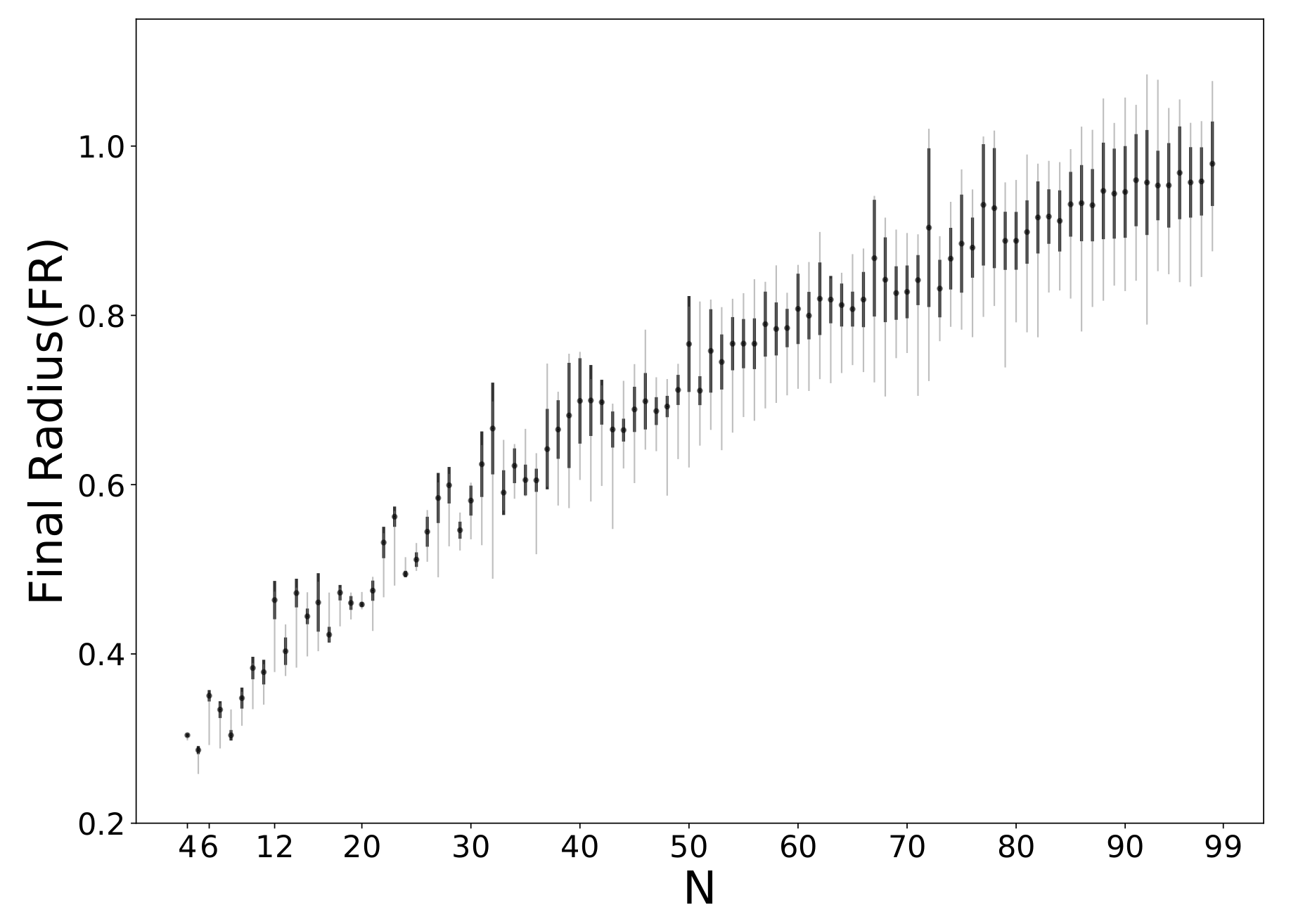}
    \caption{Maximal cage radius for different number of AUVs. One standard deviation is a thick bar, maximum and minimum values are thin bars. Mean is the dot in the middle. Experiments were repeated 100 times.}
 \label{fig:frvsn}
\end{figure}

We set the sensor volume disc radius $r_s = 0.5 / \sqrt 3$, and then examined the generated formations around a unit sphere for different numbers of AUVs.
The results can be seen in Table~\ref{tab:dists} and Figures~\ref{fig:cages} and  \ref{fig:frvsn}.

When $N=4,6,12$, the configurations are indeed close approximations of Platonic solids. 
Looking at Table~\ref{tab:dists} we see that their mean distances are very close to the analytical results, with very small standard deviations.
For $N = 5,10,20$, the formations contain faces that are not equilateral triangles and thus the distances vary much more. 
See Figure~\ref{fig:cages} for visualizations of $N=12$, $N=20$ and $N=50$.

\begin{table}[htbp]
\centering
\caption{Distributions of Maximum Inter-Agent Distance(MD) on Unit Sphere and the Corresponding Platonic Solid Edge Lengths(PS) for Comparison }
\label{tab:dists}
\begin{tabular}{@{}llll@{}}
\toprule
N  & MD Mean & MD Std. & PS Edges\\ \midrule
\textbf{4}  & \textbf{1.633}           & \textbf{0.007}                & \textbf{1.633}                             \\
5  & 1.786           & 0.061                & -                                 \\
\textbf{6}  & \textbf{1.418}           & \textbf{0.057}                & \textbf{1.414}                             \\
10 & 1.349           & 0.066                & -                                 \\
\textbf{12} & \textbf{1.058}           & \textbf{0.002}                & \textbf{1.051}                             \\
20 & 1.079           & 0.017                & -                                 \\ 
\bottomrule
\end{tabular}
\end{table}

\begin{figure}[htb!]
 \centering
 	\includegraphics[width=0.99\linewidth]{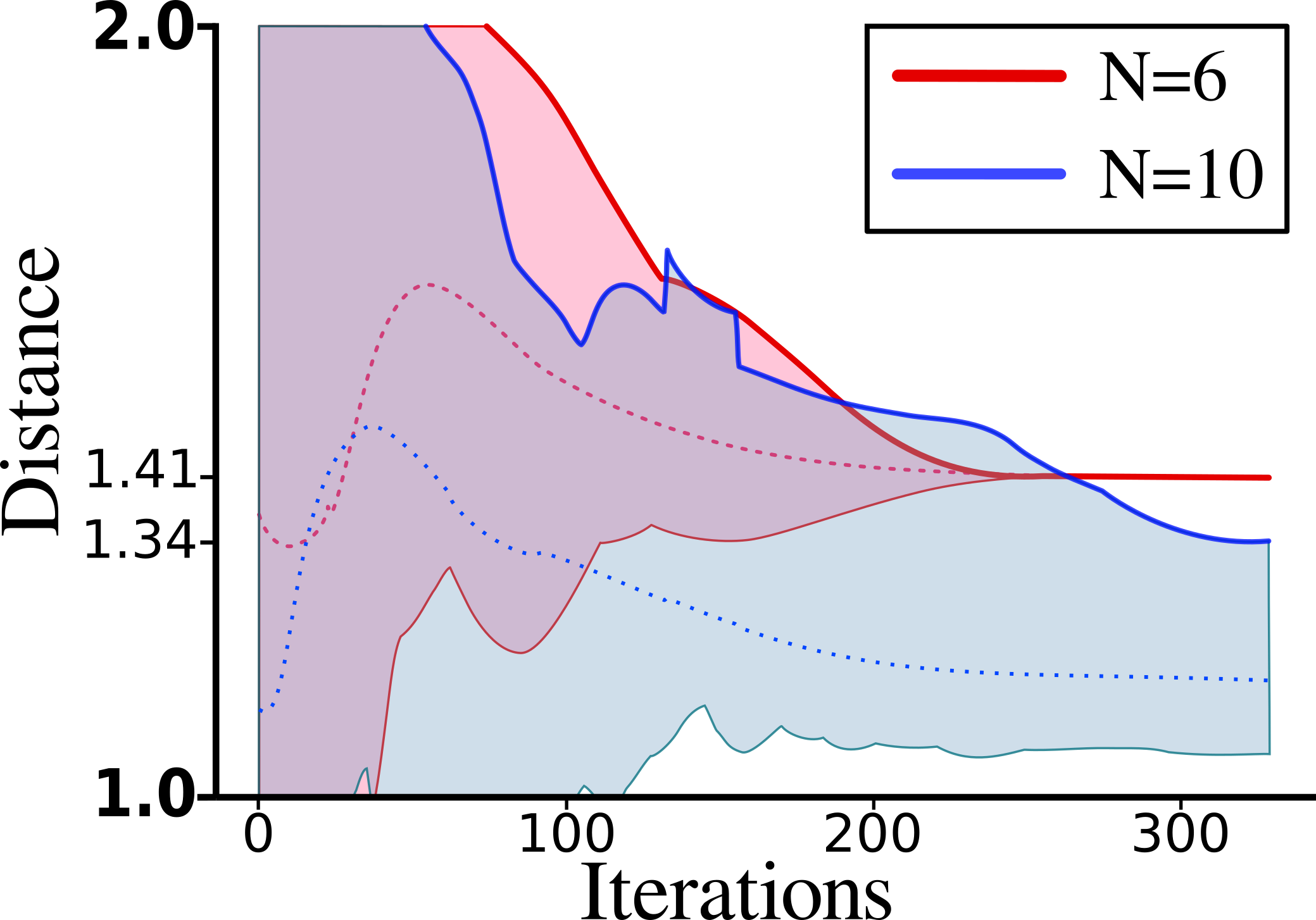}
    \caption{Convergence of maximum, mean, and minimum inter-agent distances over iterations on the unit sphere. Means are dashed, maximums are thick and minimums are thin curves.}
 \label{fig:timeseries}
\end{figure}

The fact that the distances are the same for all triangles of the Platonic solids make them very efficient for caging purposes, as the formation can be scaled up to minimize the overlap in sensor footprints. 
See Figure~\ref{fig:timeseries} to see an example with $N=6$ and $N=10$. 
When $N=6$, all distances in the cage converge to the same value. When $N=10$, all distances still converge, but not on the same value which leads to inefficient use of AUVs.
This inefficiency is caused by the need to create complete coverage of the spherical surface. 
In order to achieve this, the \textit{maximum} distance between AUVs must be limited. 
As can be seen in the figure, most AUVs are closer to each other, yet there is some that are causing the entire cage to be shrunk more.

Looking at the radius of the final cage (FR) in Figure~\ref{fig:frvsn} we see that it makes significant jumps in size for the Platonic solids. 
It is even the case that spreading five AUVs across the sphere makes a less efficient (in terms of AUV per volume caged) cage than using four. 
This result is in line with Figure~\ref{fig:timeseries}. 
Looking at Figure~\ref{fig:frvsn} we see that the irregular increase in FR continues also for larger numbers of AUVs even though the effect is not as strong as for the smaller numbers. 
From the same figure it is also evident that there is diminishing returns in terms of the caged volumes, giving us further evidence that using the obstacles in the environment might be a better idea when the contaminated volume is large.

\subsection{Limitations and Relevance to real AUV systems}
In the approach described in this paper we assume that the AUVs have sufficient capabilities in terms of localization, position control, endurance, and communication, all at a price that makes multi agent operations reasonable. 
All of these are active research areas, and it is therefore reasonable to believe that performance will improve, and costs will decrease as time progresses. 

Localization work include \cite{tan2011survey, chandrasekhar2006localization,bahr2009cooperative} where the fact that the system is comprised of many agents is used effectively to aid in individual localisation. 
Work on communication include \cite{Reed2016,korkmaz2004urban}, and finally, the price, as well as the localization and hovering problems, can be addressed by using sonar buoys, dropping additional units instead of moving the previous ones and collecting them all when the mission is completed.

In this paper we have disregarded the dynamics of AUVs for the most part. 
Many AUV models, such as the one in \cite{bhat2020}, can not hover in place while waiting for the second sighting.
Such AUVs can still be used with our method with a slight modification where AUVs unable to hover can instead move around in small circles, centered on their assigned position.
This will have the effect of the cage being broken and re-made as the AUVs align with their assigned positions with some frequency.
As long as the sensor sweeps the relevant area often enough, the cage will still be impenetrable by the caged entity due to its finite speed.
While doing such circling maneuvers, the AUVs will have to keep multiple constraints satisfied in order to avoid crashing, while still being as close to the center point as possible.
To do so, methods such as \cite{ozkahraman2020} can be utilized.

\section{CONCLUSIONS}
\label{sec:conclusions}
In this paper we have shown how a number of separate well known algorithms can be combined into an overall approach to capture entities in 3D with limited range sensors. 
These algorithms provide deterministic solutions for containing cages.

Our experiments have shown that there are certain fleet sizes that are more effective when it comes to creating capturing cages. 
In order to maximize the efficiency of a cage in terms of volume caged per vehicle used, these locally maximum fleet sizes should be used.
On the other hand, when the total volume of the cage needs to be larger than what the efficient fleet sizes can provide, larger fleets do provide more volume, with other local maximas available.
Thus, it might be in our best interest to focus on these locally maximal fleet sizes for future endeavors.

\section*{ACKNOWLEDGMENT}
This work was supported by Stiftelsen for Strategisk Forskning (SSF) through the Swedish Maritime Robotics Center (SMaRC) (IRC15-0046).

\addtolength{\textheight}{-10.8cm}   

\bibliographystyle{plain}
\bibliography{ozer}


\end{document}